\documentclass[letterpaper, 10 pt, conference]{ieeeconf}  

\IEEEoverridecommandlockouts                              

\usepackage{cite}
\usepackage{amsmath} 
\usepackage{amssymb}  
\usepackage{algorithm}
\usepackage{algorithmic}
\usepackage{graphicx}
\usepackage[pass,letterpaper]{geometry}
\usepackage{xurl}
\usepackage{multirow}
\usepackage{stfloats}
\newif\ifanonymous
\anonymousfalse 

\title{\LARGE \bf
GLidE-SLAM: GL-Accelerated Indirect-Direct Embedded SLAM
}

\ifanonymous
  \author{Anonymous Authors}
\else
  \author{Carlos A. Pinheiro de Sousa, Heiko Hamann, and Oliver Deussen%
  \thanks{This work was funded by the Deutsche Forschungsgemeinschaft (DFG, German Research Foundation) under Germany's Excellence Strategy -- EXC 2117 -- 422037984.}%
  \thanks{The authors are with the Department of Computer and Information Science, University of Konstanz, 78464 Konstanz, Germany. {\tt\small \{carlos.pinheiro-de-sousa, heiko.hamann, oliver.deussen\}@uni-konstanz.de}}%
 \thanks{{\tt\small https://github.com/capsMD}}%
  }
\fi

\begin{document}

\maketitle
\thispagestyle{empty}
\pagestyle{empty}

\begin{abstract}

With the growing demand for robotics, autonomous drones, and wearable extended reality systems, the deployment of Visual SLAM on embedded devices remains challenging. Tracking must sustain high frame rates while preserving compute resources for map extension and maintenance. This paper presents GLidE-SLAM, a monocular hybrid indirect-direct framework that addresses this by architectural separation: the system performs GPU-accelerated direct tracking on intermediate frames, while reserving the full indirect pipeline for map extension and global consistency. We leverage highly parallel image-alignment operations for pose-only estimation without depth optimization or map point creation, making the workload suitable for GPU offloading and freeing CPU resources for backend tasks. We implement the direct tracker using vendor-agnostic OpenGL ES~3.1 compute shaders, enabling deployment across a broader range of commodity embedded platforms without requiring CUDA support. To our knowledge, this is the first complete direct photometric pose estimator realized via compute shaders for embedded-class devices. Experiments on target platforms demonstrate up to 9$\times$ higher frame rates than the CPU-only baseline while maintaining trajectory accuracy and improving practical deployment across commodity resource-constrained hardware.

\end{abstract}

\section{INTRODUCTION}

Visual SLAM has been extensively studied, yet practical deployment on embedded devices remains challenging. Growing applications in robotics, autonomous drones, and wearable extended reality (XR) systems are driving demand for efficient embedded implementations. Robotics trends increasingly favor fully onboard SLAM \cite{High-Speed,ONBOARD_SLAM}; these platforms operate under strict constraints in cost, power, and form factor, where efficiency and portability matter as much as raw accuracy. 

Although recent learning-based SLAM methods \cite{GS-SLAM, Splat-SLAM, VGGT, Dust3r, mast3r} have attracted substantial attention for richer scene reconstruction and understanding, their practical requirements are mismatched to embedded deployment: they require vendor-specific, high-power GPU hardware. Classical geometric approaches and lightweight hybrid designs, therefore, remain the most practical path to real-time SLAM on small, portable, resource-constrained devices. Yet, real-time operation remains limited by the tracking front end, whose per-frame cost can starve mapping and optimization, even with multi-threaded approaches. The central challenge is therefore to reduce front-end compute while preserving robustness and map quality under tight computing and portability constraints.

Our GLidE-SLAM approach addresses this by assigning each paradigm exclusively to the regime where it is naturally strongest. A~lightweight direct tracker performs pose-only visual odometry on the majority of intermediate or \textit{tween} frames and alternates back sporadically to indirect tracking to regain stability, re-anchor the pose and extend the map structure. The relationship is symbiotic: indirect tracking constructs the map that direct tracking consumes, while frequent direct poses keep the indirect pipeline well-initialized.

Because the direct module performs pose-only estimation with no depth optimization or map point creation, its workload reduces to highly parallel local photometric operations, making it a natural fit for GPU execution. We implement a complete end-to-end direct photometric pose estimation using compute shaders. This approach not only frees CPU headroom for backend tasks that would otherwise compete with tracking on constrained platforms, but also provides a vendor-agnostic acceleration path on embedded GPUs, enabling SLAM on resource-constrained devices where it was previously prohibitive.

We provide, through this framework, at least three main contributions:

\begin{enumerate}
\item \textbf{Novel indirect-direct architectural separation.} We introduce a hybrid visual SLAM architecture that combines indirect and direct tracking according to their respective strengths: indirect tracking provides keyframe structure, map extension, and recovery, while direct photometric tracking performs efficient pose-only estimation on intermediate frames.

\item \textbf{Pose-only direct photometric tracking over a sparse map.} We formulate direct tracking as a lightweight pose-only estimator over existing sparse map points, avoiding depth optimization and map point creation during intermediate frames. This keeps the direct workload compact, parallelizable, and suitable for repeated high-rate execution.

\item \textbf{Vendor-agnostic compute-shader implementation.} To the best of our knowledge, GLidE-SLAM is the first pose-only direct photometric estimator implemented with OpenGL ES~3.1 compute shaders for embedded-class devices, supporting execution across a broader range of commodity embedded GPUs.
\end{enumerate}

\section{RELATED WORK}

\subsection{Indirect and Direct Tracking Paradigms}

The architectural separation in GLidE-SLAM builds on the complementary strengths of classical tracking methods, which fall into indirect feature-based and direct photometric paradigms. Indirect feature-based methods~\cite{MonoSLAM, ORB-SLAM, ORB-SLAM2, VINS, ORB-SLAM3} match sparse keypoints and minimize reprojection error, offering robustness to large inter-frame motion and natural support for relocalization and loop closure, but impose a high constant cost per frame through feature extraction and matching. Direct photometric methods~\cite{DTAM, DSO, LSD, LDSO}, on the other hand, minimize pixel-intensity differences between frames to achieve subpixel-accurate alignment, but are fragile under moderate motion, illumination changes, and motion blur.

\subsection{Hybrid and Semi-Direct Methods}

Semi-direct and hybrid methods aim to combine the complementary strengths of the two paradigms. LCSD-SLAM~\cite{LCSD} loosely couples direct and indirect methods~\cite{LSD, ORB-SLAM2} in parallel, but maintains two separate map representations that can drift relative to each other and increase computational overhead. H-SLAM~\cite{H-SLAM} proposes a unified inverse-depth representation shared by both direct and indirect components to eliminate dual-map drift. Closer to our approach are systems that use lightweight tracking between keyframes: SVO~\cite{SVO} refines pose via direct image alignment while jointly estimating inverse-depth for features during tracking, and OV2SLAM~\cite{OV2} performs inter-frame tracking by optical-flow feature propagation followed by Perspective-n-Point (PnP) pose refinement. In contrast, GLidE-SLAM enforces strict paradigm separation: direct tracking is restricted to camera pose-only estimation over an existing sparse map, enabling a highly parallel GPU-friendly front end for embedded systems.

\subsection{Embedded SLAM Acceleration}

Although accelerated approaches have emerged to reduce computational overhead in embedded SLAM, most systems rely on CUDA to offload front-end operations such as feature detection and tracking~\cite{FastTrack, High-Performance}, as well as back-end components including local mapping and bundle adjustment~\cite{cuVSLAM}. This vendor-specific dependence restricts deployment on commodity off-the-shelf embedded platforms. FPGA-based accelerators can achieve strong performance-per-watt for specific kernels such as ORB extraction and semi-dense tracking~\cite{ac2SLAM, eSLAM}, but require hardware-software co-design that couples algorithms to fixed platforms, further limiting flexibility and portability. Mobile AR frameworks such as ARCore and ARKit~\cite{VOMobile} are proprietary and device-specific, and thus unsuitable for reproducible research comparisons. In contrast, GLidE-SLAM designs and implements the direct photometric pose estimator natively as an OpenGL ES~3.1 compute-shader pipeline, enabling vendor-agnostic GPU acceleration across diverse embedded platforms without CUDA dependency.


\section{SYSTEM OVERVIEW}

\begin{figure}[ht]
    \centering
    \includegraphics[width=0.4\textwidth]{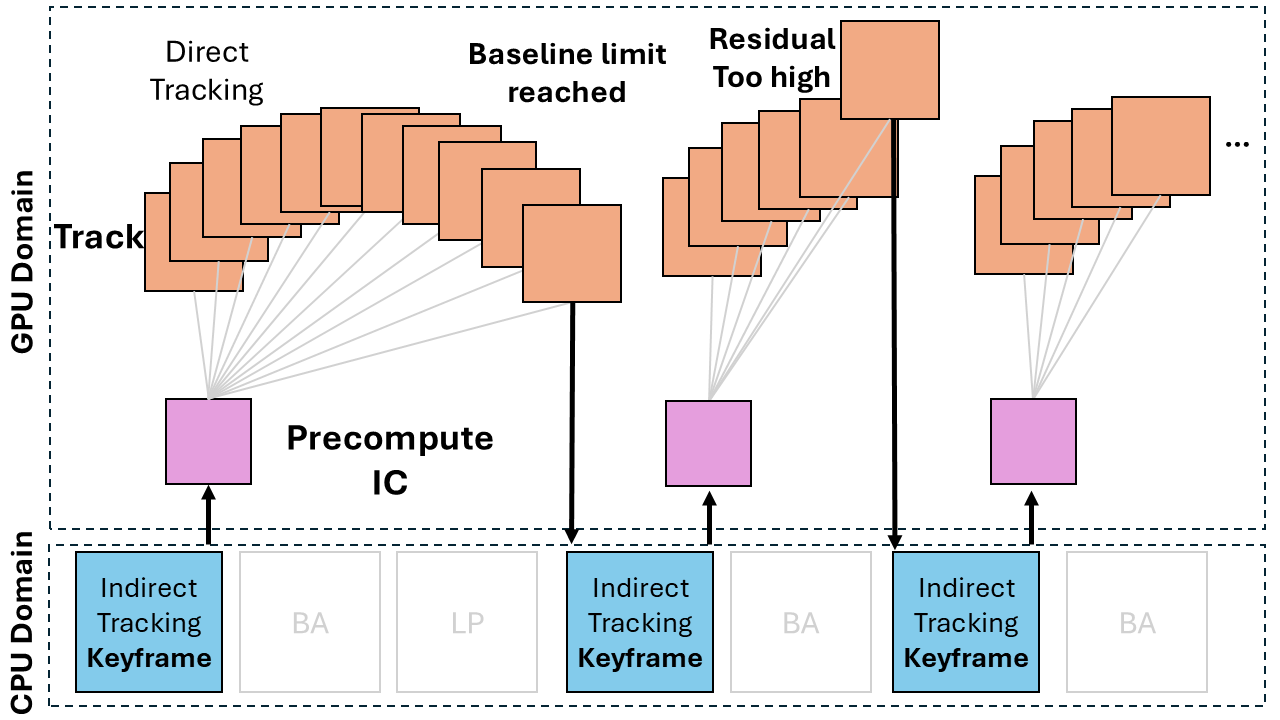}
    \caption{Windowed visual odometry schedule. Interflow GPU-CPU: \emph{Track} (orange) performs photometric alignment of sequential images to a same reference  \emph{Precompute IC} on the GPU. Indirect tracking generates keyframes, bundle adjustment (BA) and loop closure (LC) on the CPU.}
    \label{fig:WindowedVisualOdometry}
\end{figure}

GLidE-SLAM follows the standard keyframe-based organization of visual SLAM. After initialization and bootstrap map creation, the system alternates between two modes. The indirect pipeline creates \textbf{keyframes}, extends the map, and provides stable anchor poses. For the larger set of intermediate \textbf{tween frames}, GLidE-SLAM performs pose-only direct photometric tracking on the GPU. When direct alignment degrades or the baseline to the current reference exceeds a threshold, the system returns to the indirect pipeline to regain stability, re-anchor the pose, update the map, and refresh the direct reference. Fig.~\ref{fig:WindowedVisualOdometry} illustrates this schedule.

\begin{figure}[ht]
    \centering
    \includegraphics[width=0.45\textwidth]{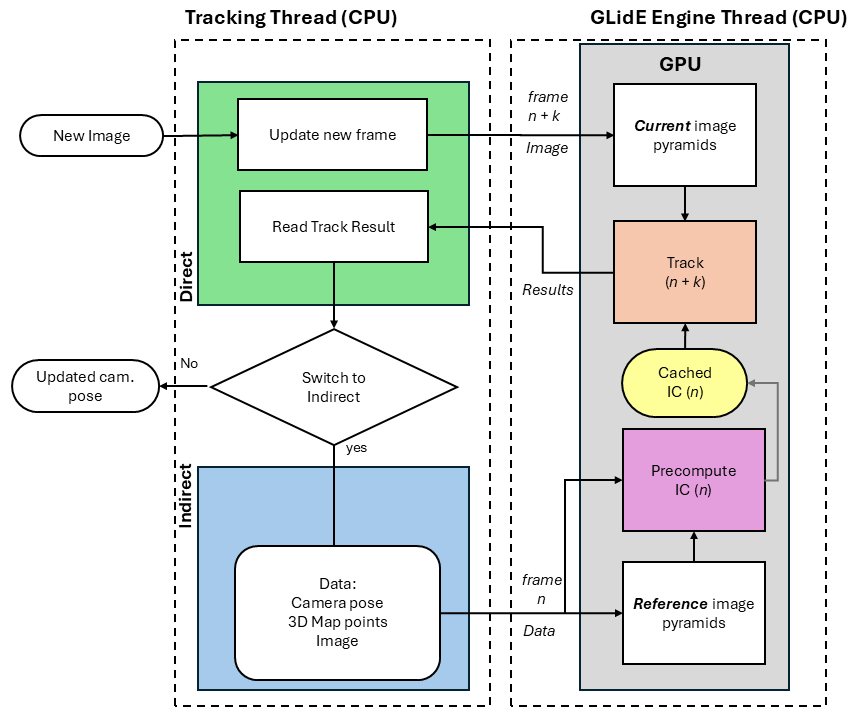}
    \caption{Direct/indirect tracking pipeline overview. GPU tracking estimates poses for current frames $n+k$ using IC precomputed from reference frame $n$.}
    \label{fig:SystemOverview}
\end{figure}

For this architecture, the indirect tracking pipeline provides the \textbf{reference} 3D map points and a corresponding reference camera pose, which serve as inputs to the GPU direct-tracking \textbf{precompute} process and cache generation. This step is performed primarily on keyframes, executed sporadically, and its results are reused within a variable-length window of consecutive frames for direct tracking. In contrast, for most sequential frames, estimates of the \textbf{current} frame's camera pose are produced by the \textbf{track} process using incoming images together with the previously cached IC reference. An overview of the two symbiotic pipelines is provided in Fig. \ref{fig:SystemOverview}, and a formal discussion is presented in \ref{sec:Direct Photometric Tracking}.

The next section formalizes the photometric alignment
objective, the inverse compositional formulation used for caching, and the main stages of our compute-shader pipeline.

\section{DIRECT PHOTOMETRIC TRACKING}
\label{sec:Direct Photometric Tracking}
\subsection{Problem Formulation}

Direct photometric tracking estimates the current camera pose $T \in \mathrm{SE}(3)$ by minimizing photometric residuals over a set of 3D map points provided by the indirect pipeline. 
Let $\{P_i\}_{i=1}^{N}$ be $N$ 3D map points. Each incoming frame is aligned to a fixed \emph{reference} frame with pose $T_0$ and reference image $I_0$. For a candidate \emph{current} pose $T$, point $P_i$ projects to
\begin{equation}
u_i^0 = \pi(T_0 P_i), \qquad u_i(T) = \pi(T P_i).
\label{eq:projections}
\end{equation}
where $\pi$ denotes the projection, $u_i^0$ denotes the projection in the reference image $I_0$, and $u_i(T)$ denotes the projection in the current image $I$.

A small $p \times p$ patch is compared around each projection. Let $\mathcal{P}=\{1,\dots,p^2\}$ index the pixels of the patch, where each $k \in \mathcal{P}$ corresponds to a fixed 2D offset $\Delta u_k$ relative to the patch center. The photometric residual for map point~$i$ and patch pixel $k$ is defined as
\begin{equation}
r_{i,k}(T) \;=\; I_0\!\left(u_i^0 + \Delta u_k\right)\;-\; I\!\left(u_i(T) + \Delta u_k\right),
\end{equation}
where intensities are sampled bilinearly.

The pose can be estimated by minimizing the sum of squared residuals for all pixels around all points:
\begin{equation}
T^* \;=\; \arg\min_{T} \sum_{i=1}^{N} \sum_{k \in \mathcal{P}} r_{i,k}(T)^2 .
\end{equation}

Optimization proceeds coarse-to-fine over the image pyramid, with the pose estimated at each coarser level used to initialize Gauss--Newton\footnote{Gauss--Newton is sufficient for this locally initialized pose-only problem with small inter-frame updates, and its fixed-iteration structure is more convenient for a compute-shader pipeline.} optimization at the next finer level, using a fixed maximum number of iterations per level. Pose increments are parameterized as a 6D twist $\delta \in \mathbb{R}^6$ and applied with the exponential map using the update per iteration:
\[
T \leftarrow T \exp(\delta).
\]

\subsection{Inverse Compositional Template Approximation}

Let $r(T)$ denote the stacked residual vector formed from
$r_{i,k}(T)$. Using Gauss--Newton, the residual is linearized around
the current estimate as
\begin{equation}
r(T\exp(\delta)) \;\approx\; r(T) + J\,\delta,
\label{eq:gn_linearization}
\end{equation}
Here, $\delta \in \mathbb{R}^6$ is the pose increment and $J$ is the
Jacobian with respect to $\delta$. Minimizing Eq.~\eqref{eq:gn_linearization}. is solved
iteratively via the normal equations
\begin{equation}
H\,\delta = b, \qquad H = J^\top J, \qquad b = -J^\top r,
\label{eq:normal_eq}
\end{equation}
In a forward-additive formulation, $J$ depends on image gradients
evaluated at warped locations in the current image $I$ and must be
recomputed at each iteration and for every new image. Inverse compositional (IC)
alignment~\cite{LKT} reduces this cost by evaluating derivatives in
the fixed reference image $I_0$, making $J$ independent of the current
pose estimate while a reference is reused. This is particularly
favorable in the tween-frame regime, where several frames are aligned
to the same reference frame.

In the IC form used here, each Jacobian row follows the standard
photometric chain rule evaluated on the reference side:
\begin{equation}
J_{i,k} \;=\;
\nabla I_0\!\left(u_i^0 + \Delta u_k\right)\,
\frac{\partial \pi}{\partial x}\,
\frac{\partial x}{\partial \delta},
\label{eq:ic_jacobian}
\end{equation}
with $x = T_0 P_i$, where $\frac{\partial \pi}{\partial x}$ is the
projection Jacobian and $\frac{\partial x}{\partial \delta}$ is the
$\mathrm{SE}(3)$ motion Jacobian. While the reference is fixed, the
Jacobian stack $J$ in Eq.~\eqref{eq:normal_eq} remains constant; only
the residual vector $r(T)$ changes with the current frame and pose
estimate. Consequently, $H$ can also be precomputed once per reference
update. Solving Eq.~\eqref{eq:normal_eq} yields the pose increment
$\delta$, which is applied via $T \leftarrow T\exp(\delta)$ to update
the current pose estimate.

\subsection{GPU Compute Shader Pipeline}
\begin{figure}[ht]
    \centering
    \includegraphics[width=0.4\textwidth]{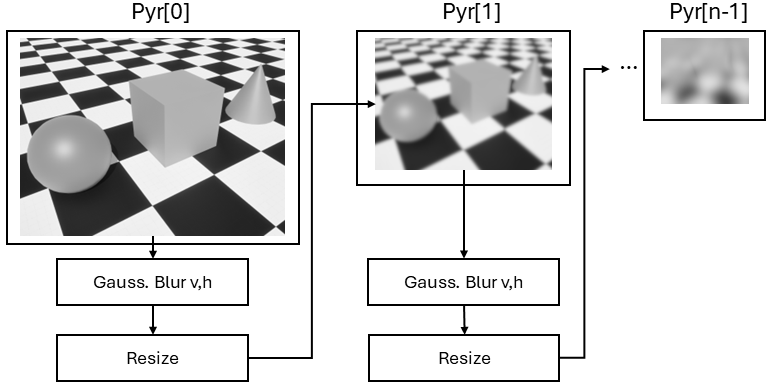}
    \caption{Multi-stage image pyramid generation.}
    \label{fig:ImagePyramids}
\end{figure}

Our GPU framework is implemented using OpenGL for Embedded Systems (OpenGL~ES)~3.1, providing portability across commodity embedded GPUs and direct support for image-processing operations. Within this framework, compute shaders perform the parallel computation required by our direct photometric pose estimator. GLidE-SLAM executes the direct photometric visual odometry pipeline end-to-end on the GPU to minimize CPU--GPU data exchange. This includes image-pyramid construction, image differentiation, tree-based reduction, and Gauss--Newton optimization. Per-point and per-patch data are stored in Shader Storage Buffer Objects (SSBOs)~\cite{SSBOs} and accessed directly by the compute-shader stages. The same framework also drives a custom lightweight 3D visualization engine. The pipeline consists of three main stages:

\textbf{Image pyramids.}
Each frame is converted to a 32-bit floating-point representation to reduce quantization error and keep downstream gradient and residual computations numerically stable. A Gaussian pyramid is then built on the GPU (Fig.~\ref{fig:ImagePyramids}). The first level ($L=0$) is full resolution, and subsequent levels are generated recursively by blurring\footnote{We use two-pass vertical/horizontal blur on a temporary texture buffer.} and downsampling the previous level.

\begin{figure}[ht]
    \centering
    \includegraphics[width=0.4\textwidth]{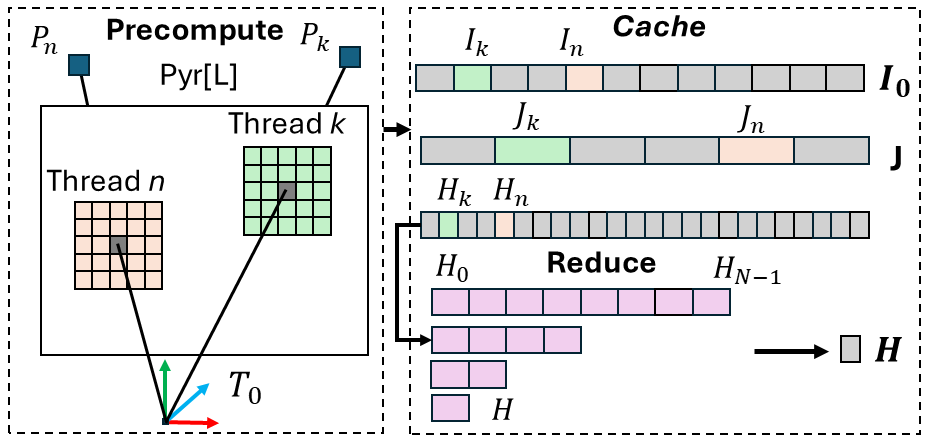}
    \caption{Per-level precompute cache storage. Per-point thread projection and buffer storage: Patch intensities $I_0$ and Jacobians $J$ are stored per patch pixel; Hessians $H$ must be reduced and stored per level.}
    \label{fig:preCompute}
\end{figure}

\textbf{Precompute Inverse Compositional (IC).} When processing a reference frame, for each of its pyramid levels, the inverse compositional cache is computed and stored in dedicated buffers ($I_0$, $J$, and $H$) shown in Fig.~\ref{fig:preCompute}. 

Each compute-shader invocation processes one map point $P_i$: it projects $P_i$ using $T_0$, samples the $p\times p$ reference patch, computes reference-side gradients, and stores the corresponding intensities and Jacobians $J_{i,k}$ from Eq.~\eqref{eq:ic_jacobian}. The Hessian contribution $H_i$ is first written per point and then reduced across points into a single per-level Hessian used during later in the track process. Since each invocation writes only to the buffer slice associated with its point index, the stage is parallel and requires no CPU-side accumulation.

\begin{figure}[ht]
    \centering
    \includegraphics[width=0.4\textwidth]{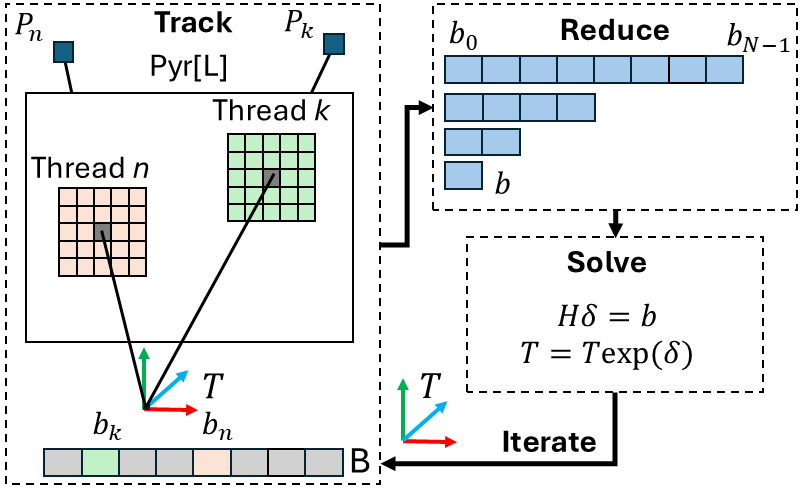}
    \caption{Per-level, per-iteration photometric alignment: per-point thread projection yields residual, which is combined with cached Jacobian to output per-point $b_i$. Posterior aggregation into per-level $b$ is then used for solving for new pose $T$.}
    \label{fig:track}
\end{figure}

\textbf{Track.}
For each current frame, tracking is performed as a three-stage compute-shader process, repeated at each pyramid level and each Gauss--Newton iteration Fig.~\ref{fig:track}.

First, one compute-shader invocation per map point projects $P_i$ using the current pose estimate $T$, samples the corresponding current-image patch, and computes photometric residuals against the cached reference patch $I_0$. For each patch pixel $k$, the residual contribution is combined with the cached Jacobian $J_{i,k}$. These per-pixel contributions are summed over the patch to produce one per-point right-hand-side vector $b_i=-\sum_k J_{i,k}^\top r_{i,k}$.

Second, a tree-reduction stage aggregates all per-point vectors $b_i$ into a single per-level 6D vector $b$, corresponding to three rotational and three translational components.

Finally, the normal equations $H\delta=b$ are solved on the GPU via Cholesky factorization to obtain the pose increment, and the pose is updated as $T \leftarrow T\exp(\delta)$. The updated pose initializes the next iteration, and the procedure is repeated across pyramid levels from coarse to fine.

\subsection{Indirect Tracking Switch}

Direct tracking is used only for tween frames and is guarded by photometric-quality and baseline criteria. A direct-tracked frame is accepted when the GPU tracker returns a valid pose and the robust photometric error remains below the switching threshold. The system returns to the indirect pipeline given any of the following conditions: if direct tracking fails, if the error exceeds the failure threshold, or if the maximum number of direct frames since the last reference is reached. Upon successful switch back to indirect tracking, the direct pipeline is refreshed with new reference data: camera pose $T_0$, input reference image and the 3D map points (see Fig.~\ref{fig:SystemOverview}); the IC precompute stage is then re-run and track is performed for next consecutive frames.

\section{EXPERIMENTAL EVALUATION}
\label{sec:experiments}

\subsection{Experimental Setup}
\label{sec:setup}

We evaluate GLidE-SLAM on three platforms comprising two embedded
devices and a development laptop: the Radxa Zero 3W (RK3566,
Mali-G52~MP2 GPU), representing a severely resource-constrained
embedded target; the NVIDIA Jetson Orin Nano (Arm Cortex-A78AE CPU,
Ampere GPU); and a development laptop (AMD Ryzen~AI~9 HX~370 CPU,
NVIDIA GeForce RTX~4060 mobile GPU). The evaluation comprises six
image sequences: three TUM RGB-D~\cite{TUM} sequences,
freiburg3\_long\_office\_household, freiburg2\_desk, and
freiburg2\_xyz, reported as tum3, tum2\_desk, and tum2\_xyz,
respectively; and three EuRoC MAV~\cite{Euroc} machine-hall sequences,
MH01, MH03, and MH05. 

All experiments use six image-pyramid levels. The direct-tracking patch size is dataset-specific: $7\times7$ for TUM RGB-D and $11\times11$ for EuRoC MAV. For each sequence, we report timing and accuracy over five runs to capture runtime performance and run-to-run variability. Rendering is disabled to avoid rendering-related bias. For this work GLidE-SLAM extends ORB-SLAM2, the evaluation focuses on the controlled question of how much tracking throughput improves on embedded devices while maintaining comparable accuracy. Broader comparisons to other SLAM frameworks would require additional porting and tuning across heterogeneous software and hardware dependencies and are outside the scope of this work.

\section{RESULTS}

We report median per-frame processing time, GPU-synchronized direct-tracking pipeline time, the percentage of frames processed by the direct tracker, and trajectory accuracy for GLidE-SLAM compared to ORB-SLAM2 on the platforms described in Sec.~\ref{sec:setup}.

\subsection{Frame Processing Performance}

Table~\ref{tab:timing} reports median per-frame processing time for ORB-SLAM2 and GLidE-SLAM on all three platforms, together with the speedup factor relative to ORB-SLAM2. GLidE-SLAM reduces per-frame processing time on most sequence-platform pairs, with the largest gains observed on sequences with high direct-tracking utilization.

\begin{table*}[t]
\vspace*{2mm}
\centering
\footnotesize
\caption{Frame Processing Time (median in ms)}
\label{tab:timing}
\begin{tabular}{l|ccc|ccc|ccc}
\hline
\multirow{2}{*}{Sequence} & \multicolumn{3}{c|}{Laptop} & \multicolumn{3}{c|}{Nano} & \multicolumn{3}{c}{Radxa} \\
& ORB & \textbf{GLidE} & \textbf{Spd.} & ORB & \textbf{GLidE} & \textbf{Spd.} & ORB & \textbf{GLidE} & \textbf{Spd.} \\
\hline
tum3        & 9.6 & 5.0 & \textbf{1.9$\times$} & 31.1 & 4.8 & \textbf{6.5$\times$} & 131.0 & 58.8 & \textbf{2.2$\times$} \\
tum2\_desk  & 9.9 & 5.1 & \textbf{1.9$\times$} & 32.8 & 3.6 & \textbf{9.0$\times$} & 132.0 & 64.3 & \textbf{2.1$\times$} \\
tum2\_xyz   & 8.5 & 3.8 & \textbf{2.2$\times$} & 29.2 & 3.1 & \textbf{9.6$\times$} & 127.0 & 43.9 & \textbf{2.9$\times$} \\
MH01        & 10.9 & 3.9 & \textbf{2.8$\times$} & 35.4 & 6.2 & \textbf{5.7$\times$} & 136.0 & 107.4 & \textbf{1.3$\times$} \\
MH03        & 9.9 & 11.9 & \textbf{0.8$\times$} & 33.4 & 32.6 & \textbf{1.0$\times$} & 135.0 & 180.5 & \textbf{0.7$\times$} \\
MH05        & 9.2 & 3.4 & \textbf{2.7$\times$} & 30.8 & 6.9 & \textbf{4.4$\times$} & 124.0 & 127.0 & \textbf{1.0$\times$} \\
\hline
\end{tabular}
\end{table*}

\subsection{GPU Timings}

The GPU timing breakdown reports synchronized compute-shader execution times for the two GPU paths used by GLidE-SLAM. Since image-pyramid construction is required before both \textit{precomputate} and \textit{track} processes, Table~\ref{tab:gpu_timing} reports \textit{Pre} as \textit{pyramid}+\textit{precompute} and \textit{Trk} as \textit{pyramid}+\textit{track}. All values are median elapsed times in milliseconds.

\begin{table}[htbp]
\centering
\footnotesize
\setlength{\tabcolsep}{6pt}
\caption{Median GPU stage timings (ms).\protect\\ Pre: pyramid+precompute; Trk: pyramid+track.}
\label{tab:gpu_timing}
\begin{tabular}{l|cc|cc|cc}
\hline
\multirow{2}{*}{Seq.} & \multicolumn{2}{c|}{Laptop} & \multicolumn{2}{c|}{Nano} & \multicolumn{2}{c}{Radxa} \\
& Pre & Trk & Pre & Trk & Pre & Trk \\
\hline
tum3       & 2.5 & 3.5 & 2.0 & 2.8 & 43.7 & 39.1 \\
tum2\_desk & 2.5 & 3.6 & 2.1 & 2.8 & 43.7 & 38.9 \\
tum2\_xyz  & 2.6 & 3.4 & 2.0 & 2.7 & 51.0 & 40.8 \\
MH01       & 1.6 & 2.6 & 2.7 & 3.9 & 69.5 & 46.4 \\
MH03       & 2.3 & 3.0 & 2.7 & 3.9 & 68.0 & 45.6 \\
MH05       & 1.1 & 1.9 & 2.8 & 3.8 & 67.3 & 45.7 \\
\hline
\end{tabular}
\end{table}

\subsection{Direct Tracking Utilization}

\begin{figure}[ht]
    \centering
    \includegraphics[width=0.4\textwidth]{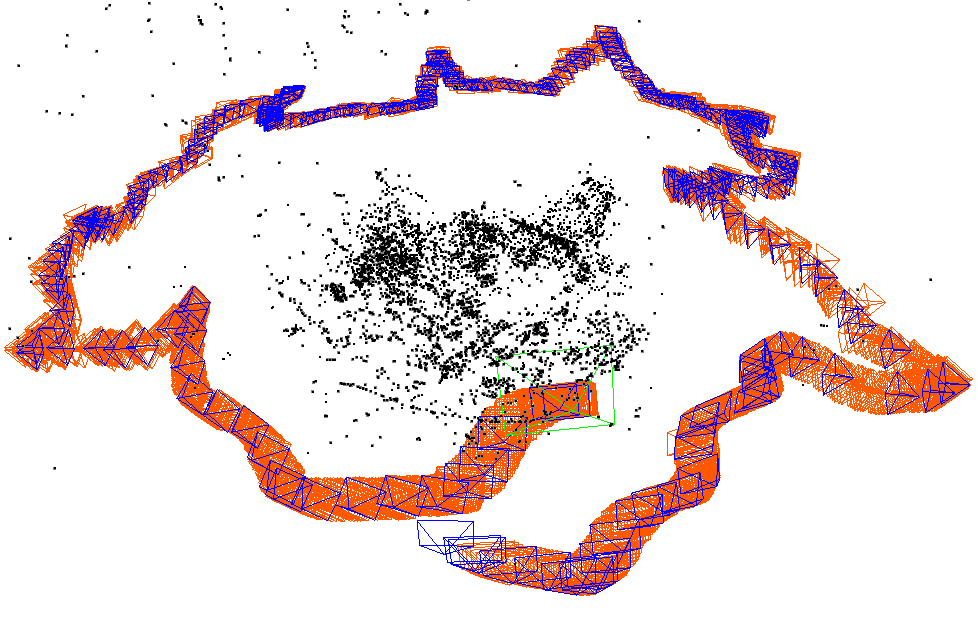}
    \caption{3D render capture of GLidE-SLAM running on fr2/desk dataset: Orange camera frames represent direct tracking, dark blue frames represent indirect tracking.}
    \label{fig:direct-indirect utilization}
\end{figure}

Higher direct-tracking utilization tends to increase speedup since direct tracking is significantly more efficient than indirect feature-based tracking. The indirect pipeline remains responsible for mapping and recovery when direct tracking is no longer reliable. This alternating relation is visible in Fig.~\ref{fig:direct-indirect utilization}, where indirect frames concentrate in regions requiring map extension or re-anchoring. Table~\ref{tab:direct_pct} reports the percentage of frames successfully processed by the direct tracker using the acceptance criterion described in Sec.~\ref{sec:Direct Photometric Tracking}.

\begin{table}[h]
\centering
\caption{Direct Tracking Success Rate (\% of frames)}
\label{tab:direct_pct}
\begin{tabular}{l|ccc}
\hline
Sequence & Laptop \textbf{GLidE} & Nano \textbf{GLidE} & Radxa \textbf{GLidE} \\
\hline
tum3      & 69.3\% & 68.5\% & 69.0\% \\
tum2\_desk & 70.1\% & 70.3\% & 69.2\% \\
tum2\_xyz  & 89.9\% & 89.8\% & 89.8\% \\
MH01      & 63.6\% & 63.9\% & 65.3\% \\
MH03      & 46.1\% & 45.0\% & 46.5\% \\
MH05      & 61.3\% & 61.2\% & 60.3\% \\
\hline
\end{tabular}
\end{table}

\subsection{Trajectory Accuracy}

Table~\ref{tab:accuracy} reports absolute trajectory error (ATE RMSE) for ORB-SLAM2 and GLidE-SLAM. Overall, GLidE-SLAM maintains accuracy across the evaluated platforms, with ATE values broadly comparable to the ORB-SLAM2 baseline.

\begin{table}[!htbp]
\centering
\footnotesize
\setlength{\tabcolsep}{4pt}
\caption{Trajectory Accuracy (ATE RMSE in meters)}
\label{tab:accuracy}
\begin{tabular}{l|cc|cc|cc}
\hline
\multirow{2}{*}{Sequence} & \multicolumn{2}{c|}{Laptop} & \multicolumn{2}{c|}{Nano} & \multicolumn{2}{c}{Radxa} \\
& ORB & GLidE & ORB & GLidE & ORB & GLidE \\
\hline
tum3      & \textbf{0.010} & 0.014 & \textbf{0.012} & 0.021 & \textbf{0.009} & 0.016 \\
tum2\_desk & \textbf{0.008} & 0.010 & 0.013 & \textbf{0.008} & \textbf{0.006} & 0.016 \\
tum2\_xyz  & \textbf{0.002} & 0.002 & \textbf{0.002} & 0.002 & \textbf{0.002} & 0.003 \\
MH01      & 0.046 & \textbf{0.044} & \textbf{0.045} & 0.048 & \textbf{0.036} & 0.042 \\
MH03      & \textbf{0.038} & 0.040 & 0.040 & \textbf{0.039} & \textbf{0.032} & 0.038 \\
MH05      & \textbf{0.056} & 0.075 & \textbf{0.051} & 0.053 & \textbf{0.037} & 0.048 \\
\hline
\end{tabular}
\end{table}


\subsection{Discussion}

GLidE-SLAM improves runtime on most sequence-platform pairs while maintaining trajectory accuracy close to ORB-SLAM2. The largest gains occur with high direct-tracking utilization, especially on TUM sequences, where many frames are processed by the GPU direct tracker. Gains are more limited on EuRoC sequences, where faster motion and stronger viewpoint changes increase inter-frame baseline, reduce direct-tracking stability, and force more frequent returns to the indirect pipeline.

All reported results use the same compute-shader configuration across platforms, including workgroup sizes and dispatch structure. Since the evaluated GPUs differ in architecture and driver behavior, the results reflect a portable but not platform-optimized configuration. Further gains may therefore be possible through device-specific shader and dispatch tuning.

\section{CONCLUSION}
We presented GLidE-SLAM, a hybrid indirect-direct SLAM architecture that performs camera pose estimation through a fast GPU-accelerated direct pipeline and invokes the indirect pipeline primarily for re-stabilization, mapping, and relocalization. Our design allows camera tracking to be highly parallelizable on embedded GPUs, achieving substantial speed gains while preserving accuracy close to the feature-based baseline. The direct pose estimator is implemented with vendor-agnostic OpenGL ES 3.1 compute shaders, supporting portability across heterogeneous embedded devices.

Future work will extend GPU acceleration to additional pipeline stages, incorporate inertial motion priors, and investigate platform-specific shader tuning. More broadly, our results highlight an opportunity beyond SLAM: embedded platforms often include capable GPUs that remain under-utilized for onboard compute, and portable GPU implementations can provide significant performance gains.



\bibliographystyle{IEEEtran}
\bibliography{root}

@article{LKT,
author = {Baker, Simon and Matthews, Iain},
year = {2004},
month = {01},
pages = {},
title = {{Lucas-Kanade} 20 Years On: A Unifying Framework Part 1: The Quantity Approximated, the Warp Update Rule, and the Gradient Descent Approximation},
journal = {International Journal of Computer Vision - IJCV}
}

@ARTICLE{MonoSLAM,
  author={Davison, Andrew J. and Reid, Ian D. and Molton, Nicholas D. and Stasse, Olivier},
  journal={IEEE Transactions on Pattern Analysis and Machine Intelligence}, 
  title={Mono{SLAM}: Real-Time Single Camera SLAM}, 
  year={2007},
  volume={29},
  number={6},
  pages={1052-1067},
  doi={10.1109/TPAMI.2007.1049}}

@ARTICLE{VINS,
  author={Qin, Tong and Li, Peiliang and Shen, Shaojie},
  journal={IEEE Transactions on Robotics}, 
  title={{VINS-Mono}: A Robust and Versatile Monocular Visual-Inertial State Estimator}, 
  year={2018},
  volume={34},
  number={4},
  pages={1004-1020},
  doi={10.1109/TRO.2018.2853729}}

@article{ORB-SLAM,
   title={{ORB-SLAM}: A Versatile and Accurate Monocular {SLAM} System},
   volume={31},
   ISSN={1941-0468},
   url={http://dx.doi.org/10.1109/TRO.2015.2463671},
   DOI={10.1109/tro.2015.2463671},
   number={5},
   journal={IEEE Transactions on Robotics},
   publisher={Institute of Electrical and Electronics Engineers (IEEE)},
   author={Mur-Artal, Raul and Montiel, J. M. M. and Tardos, Juan D.},
   year={2015},
   month=oct, pages={1147–1163} }

@article{ORB-SLAM2,
   title={ORB-SLAM2: An Open-Source SLAM System for Monocular, Stereo, and RGB-D Cameras},
   volume={33},
   ISSN={1941-0468},
   url={http://dx.doi.org/10.1109/TRO.2017.2705103},
   DOI={10.1109/tro.2017.2705103},
   number={5},
   journal={IEEE Transactions on Robotics},
   publisher={Institute of Electrical and Electronics Engineers (IEEE)},
   author={Mur-Artal, Raul and Tardos, Juan D.},
   year={2017},
   month=oct, pages={1255–1262} }

@article{ORB-SLAM3,
   title={ORB-SLAM3: An Accurate Open-Source Library for Visual, Visual–Inertial, and Multimap SLAM},
   volume={37},
   ISSN={1941-0468},
   url={http://dx.doi.org/10.1109/TRO.2021.3075644},
   DOI={10.1109/tro.2021.3075644},
   number={6},
   journal={IEEE Transactions on Robotics},
   publisher={Institute of Electrical and Electronics Engineers (IEEE)},
   author={Campos, Carlos and Elvira, Richard and Rodriguez, Juan J. Gomez and M. Montiel, Jose M. and D. Tardos, Juan},
   year={2021},
   month=dec, pages={1874–1890} }

@INPROCEEDINGS{DTAM,
  author={Newcombe, Richard A. and Lovegrove, Steven J. and Davison, Andrew J.},
  booktitle={2011 International Conference on Computer Vision}, 
  title={{DTAM}: Dense tracking and mapping in real-time}, 
  year={2011},
  volume={},
  number={},
  pages={2320-2327},
  doi={10.1109/ICCV.2011.6126513}}

@inproceedings{LSD,
 author = {J. Engel and T. Schöps and D. Cremers},
 title = {{LSD-SLAM}: Large-Scale Direct Monocular {SLAM}},
 year = {2014},
 month = {September},
 booktitle = {European Conference on Computer Vision (ECCV)},
 award = {Oral Presentation, Recipient of the ECCV 2024 Koenderink Test of Time Award},
}

@misc{DSO,
      title={Direct Sparse Odometry}, 
      author={Jakob Engel and Vladlen Koltun and Daniel Cremers},
      year={2016},
      eprint={1607.02565},
      archivePrefix={arXiv},
      primaryClass={cs.CV},
      url={https://arxiv.org/abs/1607.02565}, 
}

@INPROCEEDINGS{LDSO,
  author={Gao, Xiang and Wang, Rui and Demmel, Nikolaus and Cremers, Daniel},
  booktitle={2018 IEEE/RSJ International Conference on Intelligent Robots and Systems (IROS)}, 
  title={{LDSO}: Direct Sparse Odometry with Loop Closure}, 
  year={2018},
  volume={},
  number={},
  pages={2198-2204},
  doi={10.1109/IROS.2018.8593376}}

@INPROCEEDINGS{SVO,
  author={Forster, Christian and Pizzoli, Matia and Scaramuzza, Davide},
  booktitle={2014 IEEE International Conference on Robotics and Automation (ICRA)}, 
  title={SVO: Fast semi-direct monocular visual odometry}, 
  year={2014},
  volume={},
  number={},
  pages={15-22},
  doi={10.1109/ICRA.2014.6906584}}

@article{LCSD,
   title={Loosely-Coupled Semi-Direct Monocular SLAM},
   volume={4},
   ISSN={2377-3774},
   url={http://dx.doi.org/10.1109/LRA.2018.2889156},
   DOI={10.1109/lra.2018.2889156},
   number={2},
   journal={IEEE Robotics and Automation Letters},
   publisher={Institute of Electrical and Electronics Engineers (IEEE)},
   author={Lee, Seong Hun and Civera, Javier},
   year={2019},
   month=apr, pages={399–406} }

@misc{OV2,
      title={OV$^{2}$SLAM : A Fully Online and Versatile Visual SLAM for Real-Time Applications}, 
      author={Maxime Ferrera and Alexandre Eudes and Julien Moras and Martial Sanfourche and Guy Le Besnerais},
      year={2021},
      eprint={2102.04060},
      archivePrefix={arXiv},
      primaryClass={cs.CV},
      url={https://arxiv.org/abs/2102.04060}, 
}

@article{H-SLAM,
  title={H-SLAM: Hybrid Direct-Indirect Visual SLAM},
  author={Georges Younes and Douaa Khalil and John S. Zelek and Daniel C. Asmar},
  journal={ArXiv},
  year={2023},
  volume={abs/2306.07363},
  url={https://api.semanticscholar.org/CorpusID:259145125}
}

@misc{FastTrack,
      title={FastTrack: GPU-Accelerated Tracking for Visual SLAM}, 
      author={Kimia Khabiri and Parsa Hosseininejad and Shishir Gopinath and Karthik Dantu and Steven Y. Ko},
      year={2025},
      eprint={2509.10757},
      archivePrefix={arXiv},
      primaryClass={cs.RO},
      url={https://arxiv.org/abs/2509.10757}, 
}

@INPROCEEDINGS{High-Performance,
  author={Muzzini, Filippo and Capodieci, Nicola and Cavicchioli, Roberto and Rouxel, Benjamin},
  booktitle={2024 Design, Automation \& Test in Europe Conference \& Exhibition (DATE)}, 
  title={High-Performance Feature Extraction for GPU -Accelerated ORB-SLAMx}, 
  year={2024},
  volume={},
  number={},
  pages={1-2},
  doi={10.23919/DATE58400.2024.10546618}}

@misc{cuVSLAM,
      title={cuVSLAM: CUDA accelerated visual odometry and mapping}, 
      author={Alexander Korovko and Dmitry Slepichev and Alexander Efitorov and Aigul Dzhumamuratova and Viktor Kuznetsov and Hesam Rabeti and Joydeep Biswas and Soha Pouya},
      year={2025},
      eprint={2506.04359},
      archivePrefix={arXiv},
      primaryClass={cs.RO},
      url={https://arxiv.org/abs/2506.04359}, 
}

@INPROCEEDINGS{ac2SLAM,
  author={Wang, Cheng and Liu, Yingkun and Zuo, Kedai and Tong, Jianming and Ding, Yan and Ren, Pengju},
  booktitle={2021 International Conference on Field-Programmable Technology (ICFPT)}, 
  title={{ac2SLAM}: {FPGA} Accelerated High-Accuracy {SLAM} with Heapsort and Parallel Keypoint Extractor}, 
  year={2021},
  volume={},
  number={},
  pages={1-9},
  doi={10.1109/ICFPT52863.2021.9609808}}

@INPROCEEDINGS{eSLAM,
  author={Liu, Runze and Yang, Jianlei and Chen, Yiran and Zhao, Weisheng},
  booktitle={2019 56th ACM/IEEE Design Automation Conference (DAC)}, 
  title={eSLAM: An Energy-Efficient Accelerator for Real-Time {ORB-SLAM} on {FPGA} Platform}, 
  year={2019},
  volume={},
  number={},
  pages={1-6},
  doi={}}

@misc{VOMobile,
      title={An Empirical Evaluation of Four Off-the-Shelf Proprietary Visual-Inertial Odometry Systems}, 
      author={Jungha Kim and Minkyeong Song and Yeoeun Lee and Moonkyeong Jung and Pyojin Kim},
      year={2022},
      eprint={2207.06780},
      archivePrefix={arXiv},
      primaryClass={cs.RO},
      url={https://arxiv.org/abs/2207.06780}, 
}

@misc{mast3r,
      title={MASt3R-SLAM: Real-Time Dense SLAM with 3D Reconstruction Priors}, 
      author={Riku Murai and Eric Dexheimer and Andrew J. Davison},
      year={2025},
      eprint={2412.12392},
      archivePrefix={arXiv},
      primaryClass={cs.CV},
      url={https://arxiv.org/abs/2412.12392}, 
}

@misc{Dust3r,
      title={{DUSt3R}: Geometric 3D Vision Made Easy}, 
      author={Shuzhe Wang and Vincent Leroy and Yohann Cabon and Boris Chidlovskii and Jerome Revaud},
      year={2024},
      eprint={2312.14132},
      archivePrefix={arXiv},
      primaryClass={cs.CV},
      url={https://arxiv.org/abs/2312.14132}, 
}

@misc{VGGT,
      title={VGGT-SLAM: Dense RGB SLAM Optimized on the SL(4) Manifold}, 
      author={Dominic Maggio and Hyungtae Lim and Luca Carlone},
      year={2025},
      eprint={2505.12549},
      archivePrefix={arXiv},
      primaryClass={cs.CV},
      url={https://arxiv.org/abs/2505.12549}, 
}

@misc{GS-SLAM,
      title={GS-SLAM: Dense Visual SLAM with 3D Gaussian Splatting}, 
      author={Chi Yan and Delin Qu and Dan Xu and Bin Zhao and Zhigang Wang and Dong Wang and Xuelong Li},
      year={2024},
      eprint={2311.11700},
      archivePrefix={arXiv},
      primaryClass={cs.CV},
      url={https://arxiv.org/abs/2311.11700}, 
}

@misc{Splat-SLAM,
      title={Splat-SLAM: Globally Optimized RGB-only SLAM with 3D Gaussians}, 
      author={Erik Sandström and Keisuke Tateno and Michael Oechsle and Michael Niemeyer and Luc Van Gool and Martin R. Oswald and Federico Tombari},
      year={2024},
      eprint={2405.16544},
      archivePrefix={arXiv},
      primaryClass={cs.CV},
      url={https://arxiv.org/abs/2405.16544}, 
}

@manual{SSBOs,
  title        = {OpenGL ES 3.1 Specification},
  organization = {Khronos Group},
  edition      = {Version 3.1},
  year         = {2016},
  month        = nov,
  note         = {See Section 7.8: Shader Buffer Variables and Shader Storage Blocks},
}

@article{Euroc,
  author  = {Burri, Michael and Nikolic, Janosch and Gohl, Pascal and Schneider, Thomas and Rehder, Joern and Omari, Sammy and Achtelik, Markus W and Siegwart, Roland},
  title   = {The EuRoC micro aerial vehicle datasets},
  year    = {2016},
  doi     = {10.1177/0278364915620033},
  URL     = {http://ijr.sagepub.com/content/early/2016/01/21/0278364915620033.abstract},
  eprint  = {http://ijr.sagepub.com/content/early/2016/01/21/0278364915620033.full.pdf+html},
  journal = {The International Journal of Robotics Research}
}

@INPROCEEDINGS{TUM,
  author={Sturm, Jürgen and Engelhard, Nikolas and Endres, Felix and Burgard, Wolfram and Cremers, Daniel},
  booktitle={2012 IEEE/RSJ International Conference on Intelligent Robots and Systems}, 
  title={A benchmark for the evaluation of {RGB-D SLAM} systems}, 
  year={2012},
  volume={},
  number={},
  pages={573-580},
  doi={10.1109/IROS.2012.6385773}}

@article{ONBOARD_SLAM,
   title={Fully Onboard SLAM for Distributed Mapping With a Swarm of Nano-Drones},
   volume={11},
   ISSN={2372-2541},
   url={http://dx.doi.org/10.1109/JIOT.2024.3367451},
   DOI={10.1109/jiot.2024.3367451},
   number={20},
   journal={IEEE Internet of Things Journal},
   publisher={Institute of Electrical and Electronics Engineers (IEEE)},
   author={Friess, Carl and Niculescu, Vlad and Polonelli, Tommaso and Magno, Michele and Benini, Luca},
   year={2024},
   month=oct, pages={32363–32380} }

@ARTICLE{High-Speed,
  author={Kumar, Ashish and Park, Jaesik and Behera, Laxmidhar},
  journal={IEEE Robotics and Automation Letters}, 
  title={High-Speed Stereo Visual SLAM for Low-Powered Computing Devices}, 
  year={2024},
  volume={9},
  number={1},
  pages={499-506},
  doi={10.1109/LRA.2023.3329621}}

\end{document}